\documentclass[letterpaper, 10 pt, conference]{ieeeconf}  

\IEEEoverridecommandlockouts 

\usepackage{tabularx}
\usepackage{multirow}
\usepackage{amssymb} 
\usepackage{lipsum} 
\usepackage{amssymb}
\usepackage{pifont}
\usepackage{utfsym}
\usepackage{threeparttable}
\usepackage{cite}
\usepackage{amsmath,amssymb,amsfonts}
\usepackage{algorithmic}
\usepackage{graphicx}
\usepackage{hyperref}
\usepackage{textcomp}
\usepackage{url}
\usepackage{booktabs}
\usepackage{xcolor}

\DeclareUnicodeCharacter{221A}{$\sqrt{}$}

\title{\LARGE \bf
CARLA-Loc: Synthetic SLAM Dataset with Full-stack Sensor Setup in Challenging Weather and Dynamic Environments
}

\author{Yuhang Han$^{1}$, Zhengtao Liu$^{1}$, Shuo Sun$^{1}$, Dongen Li$^{1}$, Jiawei Sun$^{1}$, Chengran Yuan$^{1}$ and Marcelo H. Ang Jr.$^{1}$
\thanks{$^{1}$Authors are with Advanced Robotics Centre (ARC), National University of Singapore, 117575, Singapore.
{\tt\small \{yuhang\_han\normalfont, \tt\small zhengtao\_l\normalfont, \tt\small shuo.sun\normalfont, \tt\small li.dongen\normalfont, \tt\small sunjiawei\normalfont, \tt\small chengran.yuan\}@u.nus.edu\normalfont, \tt\small mpeang@nus.edu.sg}}%
}

\begin{document}

\maketitle

\begin{abstract}

The robustness of SLAM (Simultaneous Localization and Mapping) algorithms under challenging environmental conditions is critical for the success of autonomous driving. However, the real-world impact of such conditions remains largely unexplored due to the difficulty of altering environmental parameters in a controlled manner. To address this, we introduce CARLA-Loc, a synthetic dataset designed for challenging and dynamic environments, created using the CARLA simulator. Our dataset integrates a variety of sensors, including cameras, event cameras, LiDAR, radar, and IMU, etc. with tuned parameters and modifications to ensure the realism of the generated data. CARLA-Loc comprises 7 maps and 42 sequences, each varying in dynamics and weather conditions. Additionally, a pipeline script is provided that allows users to generate custom sequences conveniently. We evaluated 5 visual-based and 4 LiDAR-based SLAM algorithms across different sequences, analyzing how various challenging environmental factors influence localization accuracy. Our findings demonstrate the utility of the CARLA-Loc dataset in validating the efficacy of SLAM algorithms under diverse conditions.
CARLA-Loc dataset is made publicly available at \color{blue}{\url{https://yuhang1008.github.io/CARLA-Loc_page/}}.
\end{abstract}

\section{Introduction}
Performing accurate and robust localization and mapping in  highly dynamic and challenging weather environments is challenging for autonomous driving applications.

Currently, both LiDAR-based and visual-based SLAM methods rely on the frame-to-frame or frame-to-map matching. 
The presence of dynamic objects such as vehicles, pedestrians
, and cyclists, introduce mismatches of features in consecutive frames. Two consequences are typically brought by this effect. Firstly, the overall localization accuracy is directly affected by mismatching of dynamic object features~\cite{9636814}. Outlier-removal algorithms like random sample consensus (RANSAC) ~\cite{10.1145/358669.358692}
or using the robust loss functions 
cannot identify outliers correctly when majority of features come from dynamic objects. Secondly, the overall mapping quality is vulnerable to the accumulative noisy features from moving objects, such as \textit{ghost trails}~\cite{9197168}. 
Similar mechanism to the quality of the sparse mapping and dense reconstruction for visual SLAM~\cite
{10.1145/3177853}.

\begin{figure}[t]
    \includegraphics[width = 8.8cm]{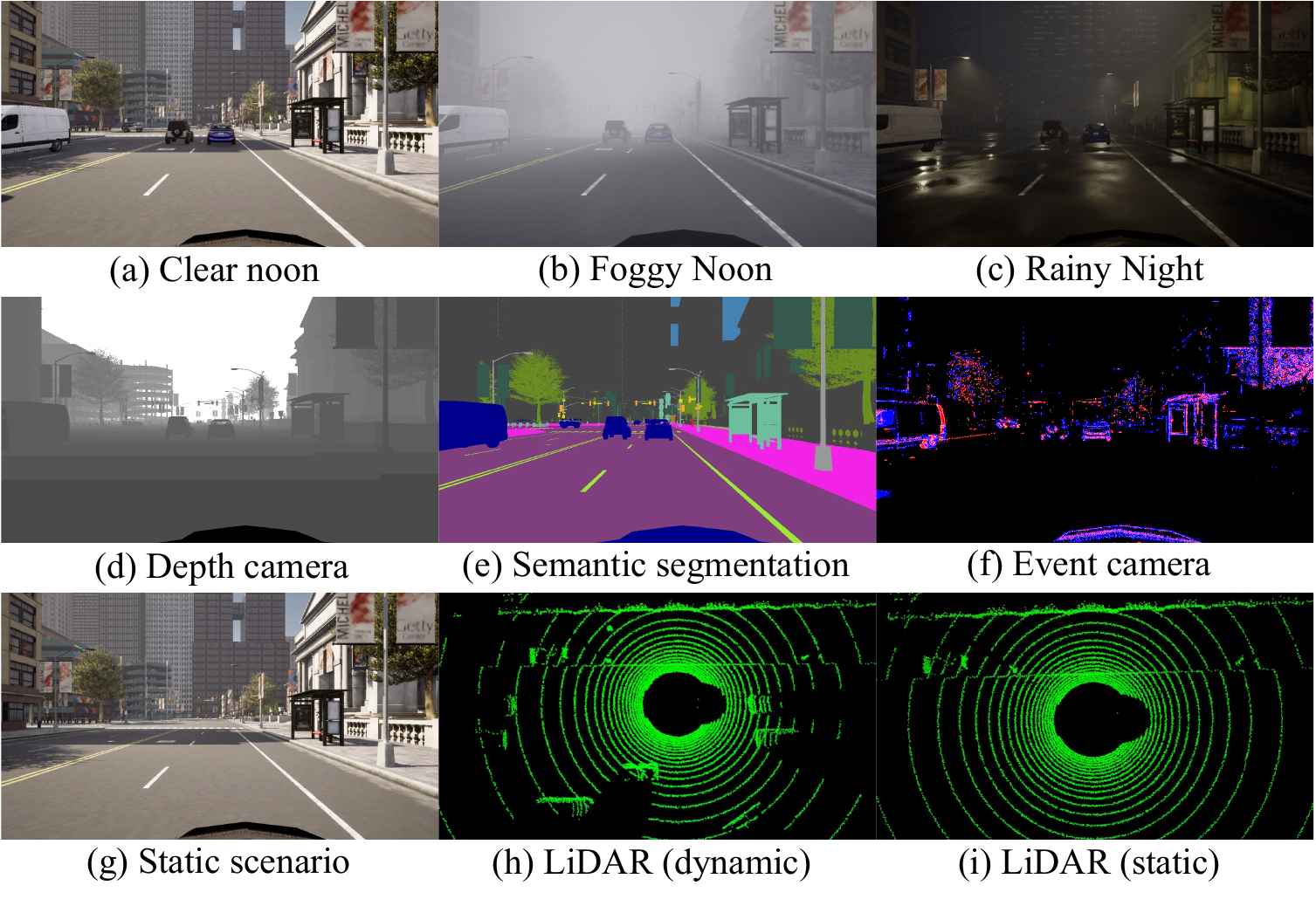}
    \caption{Overview of CARLA-Loc dataset. (a)-(c) show 3 preset weathers (clear noon, foggy noon, rainy night) for each map. (d)-(f) are images captured from depth camera, segmentation and event camera, respectively. (g) is the image from the static environment, all data are collected in the static setup as well. (h) is one scan of LiDAR in the dynamic environment, while (i) is the point cloud from static setup. }
    \label{fig:intro}
\end{figure}

A potential strategy to reduce the inaccuracies caused by precense of dynamic objects involves incorporating is sensors into the SLAM system. Inertial-added SLAM methods take advantage of the transformation residual obtained from IMU pre-integration~\cite{7557075} and integrate this data into the overall cost functions, together with feature-based errors. This principle underpins both Visual-Inertial Odometry (VIO) and LiDAR-Inertial Odometry (LIO).  Nevertheless, it is important to acknowledge that inertial-aided odometry may still face reliability issues when the observations from cameras or LiDAR degrade significantly.

Another factor poses unavoidable challenge is the poor weather. In conditions such as heavy rain, haze, and environments abundant in reflective objects, the quality of imaging significantly deteriorates, impairing the accuracy of visual odometry. Moreover, haze and rainfall result in the degradation of LiDAR scanning, yielding exceedingly noisy and irregular point clouds, which further compromise the robustness of the LiDAR-based odometry~\cite{10160874}.

One of major disadvantages of real-world datasets is ensuring consistent data collection across different environmental conditions while maintaining identical ego motion for quantitative SLAM performance evaluations. Additionally, existing simulation datasets often fail to cover both weather variations and dynamic scenes comprehensively, and their sensor configurations may not be extensive. To address these gaps and facilitate a comparative analysis of SLAM performance under diverse weather conditions and in the presence of dynamic objects, we introduce CARLA-Loc. This synthetic dataset, designed for SLAM performance evaluation in challenging scenarios, incorporates multiple built-in sensors and is developed using the CARLA simulator~\cite{pmlr-v78-dosovitskiy17a}. Within each map, CARLA-Loc provides multiple twin sequences, introducing varying weather and dynamic conditions while ensuring identical ego trajectories. This design enables a direct evaluation on the robustness of SLAM algorithms. Our experiments with various state-of-the-art visual-based and LiDAR-based SLAM algorithms on this CARLA-Loc have yielded quantified localization and mapping results, offering a comprehensive framework for performance comparison.

The main contributions of this work are:
\begin{itemize}

\item Rather than relying on the default sensor configurations provided by the CARLA simulator, we meticulously customized each sensor with finely tuned parameters to emulate commonly used real-world sensors. The IMU data was completely overhauled to include a more realistic error model, and we introduced a point cloud distortion effect to enhance the authenticity of the data further.

\item By leveraging the recording functionality of the CARLA simulator, our dataset ensures identical ego motion across different environmental settings within each map. This consistency is crucial for facilitating qualitative comparisons and advancing the design of more robust localization algorithms.

\item CARLA-Loc is more than just a dataset. It offers a complete pipeline that enables users to record their sequences according to specific design requirements, using our finely tuned and modified sensors. Additionally, we provide the script for converting complete raw data to ROS bags, offering a more convenient and more efficient solution than the original CARLA ROS bridge for generating ROS bags.

\end{itemize}
\section{Related Work}
\subsection{SLAM in Dynamic Environments}
In the realm of visual SLAM, DynaSLAM~\cite{8421015} stands out as an early work that integrates Mask R-CNN~\cite{He_2017_ICCV} with multi-view geometry constraints, demonstrating improved localization accuracy over ORB-SLAM2~\cite{7946260}. Following this, DynaSLAM II~\cite{9385844} introduces a tracking module, incorporating both velocity and reprojection error from dynamic objects into the bundle adjustment. This principle, along with the use of visual object detection or semantic segmentation, has been applied in similar works~\cite{8593691,9561452,9830851}. However, the effectiveness of these methods depends on the efficiency and accuracy of the underlying deep learning network. More recently, Seungwon\textit{ et al.}~\cite{9870851} proposed DynaVINS, which foregoes the use of a perception module. Instead, it replaces the Huber function with a proposed function that includes a weight factor to adjust the importance of visual reprojection error, offering a novel approach to dynamic object feature filtering for inertial-aided SLAM methods.

Similarly, in LiDAR SLAM under dynamic conditions, deep learning techniques have been extensively utilized. A common approach in various studies~\cite{8967704, 9526756, rs13183651} involves converting LiDAR scans into range images and then employing a semantic segmentation network to classify objects. The next step is to verify the consistency of each object patch to determine its dynamic status. Moreover, 3D point cloud object detection methods are also combined in LiDAR SLAM methods to reduce the impact of moving objects. For instance, Chen\textit{ et al.}\cite{9921969} implemented PointPillars\cite{Lang_2019_CVPR} as a detection front-end to identify objects and remove the points within them. Additionally, IMU information is used to provide a motion prior for LiDAR SLAM. RF-LIO~\cite{9636624} exemplifies this by using IMU preintegration for ego-motion estimation and assessing the visibility change in resolution of range images to detect dynamic objects.

\subsection{Existing Datasets for SLAM Evaluation}




\begin{table*}[htb]
\caption{Comparison of existing dataset with ours}
\label{tab:review}
\resizebox{\textwidth}{!}{%
\begin{threeparttable}
\def\arraystretch{1}
\begin{tabular}{|l|l|l|l|l|l|l|l|}
\hline
name                                             & year & camera(s)                                                                                                                                          & LiDAR                         & Radar          & IMU            & GNSS           & Groud truth                                                                                                \\ \hline \hline
KITTI~\cite{geiger2013vision}                    & 2013 & RGB/grayscale stereo 1382$\times$512 @ 10 Hz                                                                                                            & Velodyne   HDL-64E            & \usym{2717}     & \usym{2713} & \usym{2713} & \begin{tabular}[c]{@{}l@{}}fused pose from OXTS RT 3003 \\ @ 10 Hz, acc. \textless{} 10 cm\end{tabular}    \\ \hline
Malaga Urban~\cite{doi:10.1177/0278364913507326} & 2014 & RGB stereo 1024$\times$768 @ 20 Hz                                                                                                                      & \usym{2717}                    & \usym{2717}     & \usym{2713} & \usym{2713} & GPS @ 1 Hz, low acc                                                                                        \\ \hline
Umich NCLT~\cite{doi:10.1177/0278364915614638}   & 2015 & RGB Omnidirectional @ 16 Hz                                                                                                                        & Velodyne HDL-32E              & \usym{2717}     & \usym{2713} & \usym{2713} & \begin{tabular}[c]{@{}l@{}}fused GPS/IMU/laser pose\\  @150 Hz, acc$\approx$ 10 cm\end{tabular}            \\ \hline
EuRoC MAV~\cite{doi:10.1177/0278364915620033}    & 2016 & grayscale stereo 752$\times$480 @ 20 Hz                                                                                                                 & \usym{2717}                    & \usym{2717}     & \usym{2713} & \usym{2717}     & \begin{tabular}[c]{@{}l@{}}motion capture pose @ 100Hz, \\ acc $\approx$ 1 mm\end{tabular}                 \\ \hline
Zurich Urban~\cite{doi:10.1177/0278364917702237} & 2017 & RGB monocular 1920$\times$1080 @ 30 Hz                                                                                                                  & \usym{2717}                    & \usym{2717}     & \usym{2713} & \usym{2713} & Pix4D visual pose, unkonwn acc                                                                             \\ \hline
TUM VI~\cite{8593419}                            & 2018 & grayscale stereo 1024$\times$1024 @ 20 Hz                                                                                                               & \usym{2717}                    & \usym{2717}     & \usym{2713} & \usym{2717}     & \begin{tabular}[c]{@{}l@{}}partial motion capture pose @ 120Hz, \\ marker pos acc$\approx$1mm\end{tabular} \\ \hline
UrbanLoco~\cite{9196526}                         & 2020 & 6 x RGB Monocular 1920$\times$1200 @ 10 Hz                                                                                                              & RS-LiDAR-32                   & \usym{2717}     & \usym{2713} & \usym{2713} & \begin{tabular}[c]{@{}l@{}}fused from RTK and IMU @ 1Hz,\\ acc\textless{}12cm\end{tabular}                 \\ \hline
VIODE$^1$~\cite{9351597}                             & 2021 & RGB/SS$^*$ stereo 752$\times$480 @ 20 Hz                                                                                                                    & \usym{2717}                    & \usym{2717}     & \usym{2713} & \usym{2717}     & simulation @ 200Hz                                                                                         \\ \hline
TUM VIE~\cite{9636728}                           & 2021 & \begin{tabular}[c]{@{}l@{}}grayscale stereo 1024$\times$1024 @ 20 Hz, \\ event stereo 1024$\times$1024\end{tabular}                                          & \usym{2717}                    & \usym{2717}     & \usym{2713} & \usym{2717}     & \begin{tabular}[c]{@{}l@{}}motion capture pose @ 120Hz, \\ acc\textless {} 1 mm\end{tabular}               \\ \hline
IBISCape$^1$~\cite{soliman2022ibiscape}              & 2022 & \begin{tabular}[c]{@{}l@{}}RGB stereo 1024$\times$1024 @ 20 Hz, \\ depth/ss monocular 1024$\times$1024  @ 20 Hz,\\  event monocular 1024$\times$1024\end{tabular} & \begin{tabular}[c]{@{}l@{}} 64-channel LiDAR \\ in simulation @ 20 Hz \end{tabular} & \usym{2717}     & \usym{2713} & \usym{2713} & simulation @ 200Hz                                                                                         \\ \hline
\textbf{CARLA-Loc (ours)}$^1$                                 & 2023 & \begin{tabular}[c]{@{}l@{}}RGB/depth/SS$^*$ stereo 1280$\times$720 @ 20 Hz, \\ event stereo 1280$\times$720\end{tabular}                                         & \begin{tabular}[c]{@{}l@{}} 32-channel LiDAR \& SS$^*$ \\ in simulation @ 20 Hz \end{tabular} & \usym{2713} & \usym{2713} & \usym{2713} & simulation @ 200Hz                                                                                         \\ \hline
\end{tabular}%
\begin{tablenotes}
  \small
  \item[1] Dataset created in simulation.
  \item[*] SS stands for semantic segmentation.
\end{tablenotes}
\end{threeparttable}
}
\end{table*}

The details of prominent datasets for localization benchmarks over the past decade are summarized in Table~\ref{tab:review}. A significant challenge in creating real-time datasets, especially in outdoor settings, is obtaining accurate ground truth poses. Projects such as Malaga Urban~\cite{doi:10.1177/0278364913507326} and Zurich Urban~\cite{doi:10.1177/0278364917702237} have relied on GPS or visual methods to generate ground truth data, but these approaches offer limited accuracy. Given the impracticality of deploying motion capture systems across large areas, sensor fusion has emerged as the most dependable technique for establishing ground truth~\cite{geiger2013vision,doi:10.1177/0278364915614638,9196526}.

With the development of multi sensor SLAM algorithms and the need of more robust localization performance, more challenging datasets are proposed in recent years. For instance, datasets like EuRoC MAV~\cite{doi:10.1177/0278364915620033} and VIODE~\cite{9351597} are designed with light-weight carriers, such as UAVs, that exhibit rapid motion changes. Additionally, environmental factors contribute to the challenge by generating imperfect and noisy data. The UrbanLoco~\cite{9196526} dataset, recorded in dynamic, large-scale urban environments with trajectories exceeding 40 kilometers, exemplifies this.

While real-world datasets offer extensive coverage, they are not without their limitations. First, accurately capturing ground truth poses requires complex methods. Second, the impact of dynamic objects and environmental factors on localization accuracy cannot be quantitatively assessed. However, advancements in computer simulation engine technology have facilitated the creation of simulation-based datasets. For example, VIODE~\cite{9351597} utilizes AirSim~\cite{10.1007/978-3-319-67361-5_40} to replicate the same ego trajectories under varying dynamics. Similarly, IBISCape~\cite{soliman2022ibiscape}, akin to our work, uses the CARLA~\cite{pmlr-v78-dosovitskiy17a} simulator to generate synthetic data. Our research builds upon these prior efforts by improving ego motion control, offering more realistic sensor configurations, and providing a pipeline for users to create customized sequences.

\section{The CARLA-Loc Dataset}

The CARLA-Loc Dataset is built upon CARLA\footnote{\href{https://carla.org/}{https://carla.org/}}~\cite{pmlr-v78-dosovitskiy17a} simulator. We recorded various sequences in 7 different maps to facilitate quantitative comparisons, including different weather and dynamic conditions. These configurations of our dataset offer enhanced diversity and complexity compared to existing works (Tab.~\ref{tab:review_condition}).

\begin{table}[htb]
\caption{Comparison of enviromental conditions to existing dataset.}
\label{tab:review_condition}
\resizebox{0.5\textwidth}{!}{%
\begin{threeparttable}
\def\arraystretch{1}
\begin{tabular}{|l|l|l|l|}
\hline
name                                             & dynamic level & weather condition & time of day   \\ \hline \hline
KITTI~\cite{geiger2013vision}                    & low           & clear             & daytime       \\ \hline
Malaga Urban~\cite{doi:10.1177/0278364913507326} & mid           & clear             & daytime       \\ \hline
Umich NCLT~\cite{doi:10.1177/0278364915614638}   & low           & clear             & daytime       \\ \hline
EuRoC MAV~\cite{doi:10.1177/0278364915620033}    & N/A (indoors) & N/A (indoors)     & N/A (indoors) \\ \hline
Zurich Urban~\cite{doi:10.1177/0278364917702237} & low           & clear             & daytime       \\ \hline
TUM VI~\cite{8593419}                            & low           & clear             & daytime       \\ \hline
UrbanLoco~\cite{9196526}                         & diverse$^*$       & clear             & daytime       \\ \hline
VIODE$^1$~\cite{9351597}                             & high          & clear             & daytime/night \\ \hline
TUM VIE~\cite{9636728}                           & N/A (indoors) & N/A (indoors)     & N/A (indoors) \\ \hline
IBISCape$^1$~\cite{soliman2022ibiscape}              & diverse$^*$       & dynamic           & daytime/night \\ \hline
\textbf{CARLA-Loc (ours)} $^1$                              & diverse$^*$       & dynamic           & daytime/night \\ \hline
\end{tabular}%
\begin{tablenotes}
  \small
  \item[1] Dataset created in simulation.
  \item[*] Diverse stands for different dynamic level while ensuring same ego motion, this is typically achieved only in the simulation.
\end{tablenotes}
\vspace{-0.5cm}
\end{threeparttable}
}
\end{table}

\subsection{Dataset description}

There are 7 maps used in the dataset that are selected from CARLA, including Town01, Town03 to Town07, and Town10. These maps encompass a variety of environments such as cities, rural areas, and highways to enhance diversity. Each map is set in both dynamic and static levels of activity and is recorded under 3 pre-set weather conditions (Clear Noon, Foggy Noon, Rainy Night). Therefore, each map offers 6 sequences, resulting in a total of 42 sequences available for testing within the dataset. Detailed information about the maps and weather conditions can be found in Table~\ref{tab:summary}.

\begin{table}[h]
\caption{Parameter settings for each map in CARLA-Loc}
\label{tab:summary}
\resizebox{0.49\textwidth}{!}{%
\begin{tabular}{|c|c|c|c|c|c|}
\hline
\begin{tabular}[c]{@{}c@{}}map \\ sequence\end{tabular} & environment     & \begin{tabular}[c]{@{}c@{}}\# of \\ vehicles\end{tabular} & \begin{tabular}[c]{@{}c@{}}\# of \\ walkers\end{tabular} & \begin{tabular}[c]{@{}c@{}}total \\ lenth (m)\end{tabular} & \begin{tabular}[c]{@{}c@{}}loop \\ closure\end{tabular} \\
\hline \hline
01                                              & small town      & 102                                                       & 30                                                       & 678.20                                                     & $\times$                                                       \\ \hline

02                                              & mid city        & 237                                                       & 34                                                       & 969.64                                                     & $\checkmark$                                                       \\ \hline
03                                              & highway         & 296                                                       & 34                                                       & 2108.07                                                    & $\checkmark$                                                       \\ \hline
04                                              & mid city        & 241                                                       & 44                                                       & 473.40                                                     & $\times$                                                       \\ \hline
05                                              & rural area      & 218                                                       & 50                                                       & 265.96                                                     & $\times$                                                       \\ \hline
06                                              & arable land     & 91                                                        & 36                                                       & 747.23                                                     & $\times$                                                       \\ \hline
07                                              & skyscraper city & 108                                                       & 56                                                       & 1114.91                                                    & $\checkmark$\\
\hline
\end{tabular}
}
\end{table}

To enhance the realism of vehicle behaviors in the dataset, including lane changing, traffic jams, and overtaking, velocity of all vehicles are adjusted to vary between $80\%$ and $120\%$ of the default value on each map.


\subsection{Dataset generation}

\begin{figure}[h]
    \includegraphics[width = 8.5cm]{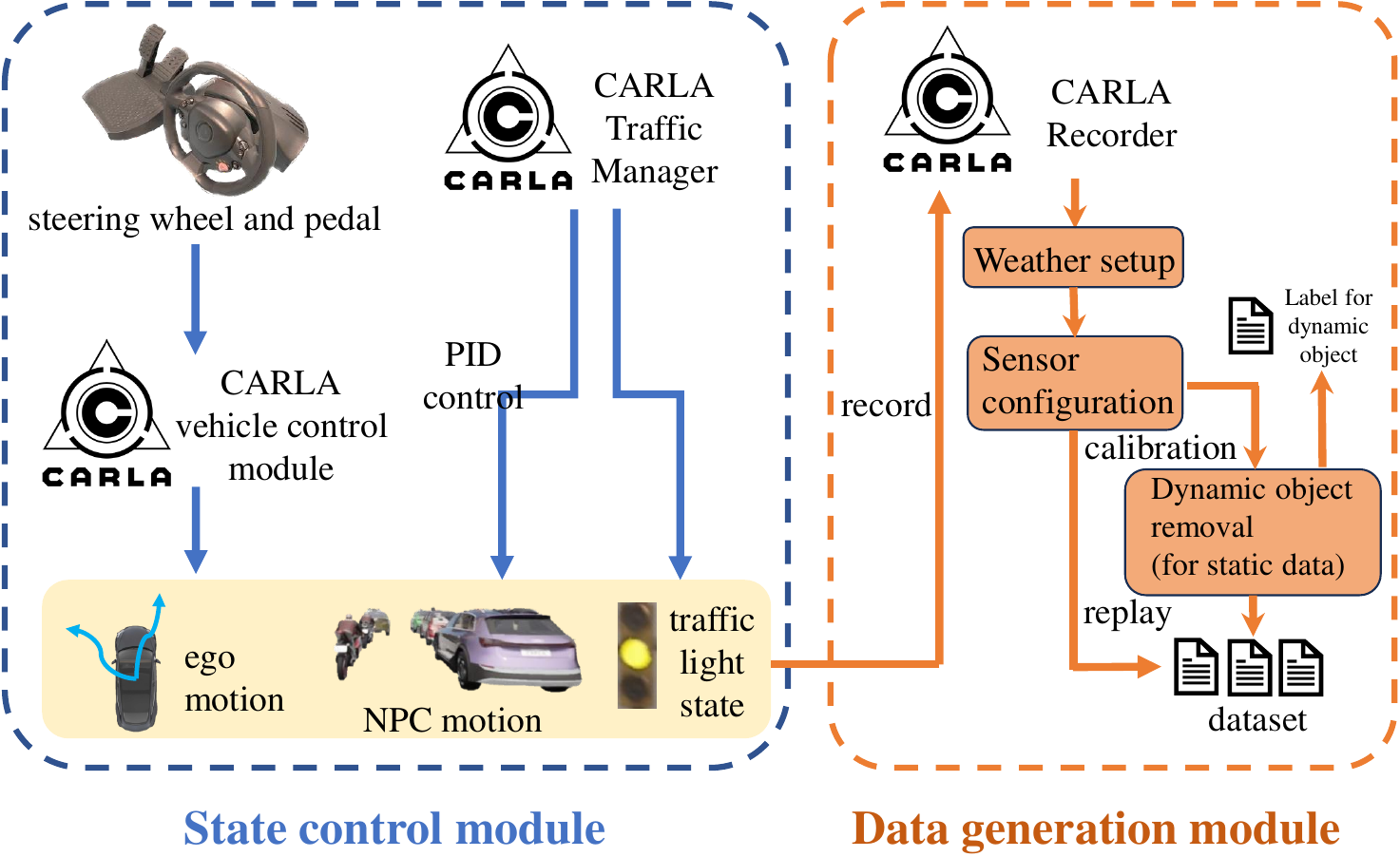}
    \caption{The pipeline to generate CARLA-Loc dataset. The ego motion is controlled through steering wheel and pedal manually, while the motion of other vehicles and pedestrians are automatically controlled by Traffic Manager of CARLA. The object motions and traffic light states are recorded by CARLA Recorder. The recordings are replayed in the given conditions to generate the diverse final dataset.  }
    \label{fig:data generation}
\end{figure}

We leverage the recorder function to create dataset under divers conditions (weather, dynamic level) in each map. The entire process for dataset generation is illustrated in Fig.~\ref{fig:data generation}.

Initially, the states of all objects and traffic lights are captured using the CARLA recorder. One observation is that the default vehicle control in CARLA is managed by a simple PID controller. Relying solely on PID control leads to unrealistic and inauthentic movements, especially evident in the IMU data during vehicle start-up and turning, where intermittent and discontinuous acceleration and angular velocity data are noticeable. To mitigate this, ego vehicle is controlled manually using a steering wheel and pedals, while CARLA automatically manages other objects, given that inertial data for these entities are not required.

Before replaying the recording, several attributes of the simulated world are customized according to the specification of different sequences, including weather and numbers of agents and light states of them. The CARLA world is set to synchronous mode to promise the strict timestamp synchronization as well as integrity of the data flow.

CARLA-Loc provides users with scripts for easy customization and dataset creation, allowing for straightforward adjustments to environmental settings and agent configurations. Recording the ego vehicle's motion once automatically generates all sub-sequences. Additionally, the toolkit to generate ros bags operates independently of the CARLA ROS bridge, enabling convenient and efficient bag creation.

\subsection{Sensor setup}

\begin{figure}[h]
    \includegraphics[width = 8.5cm]{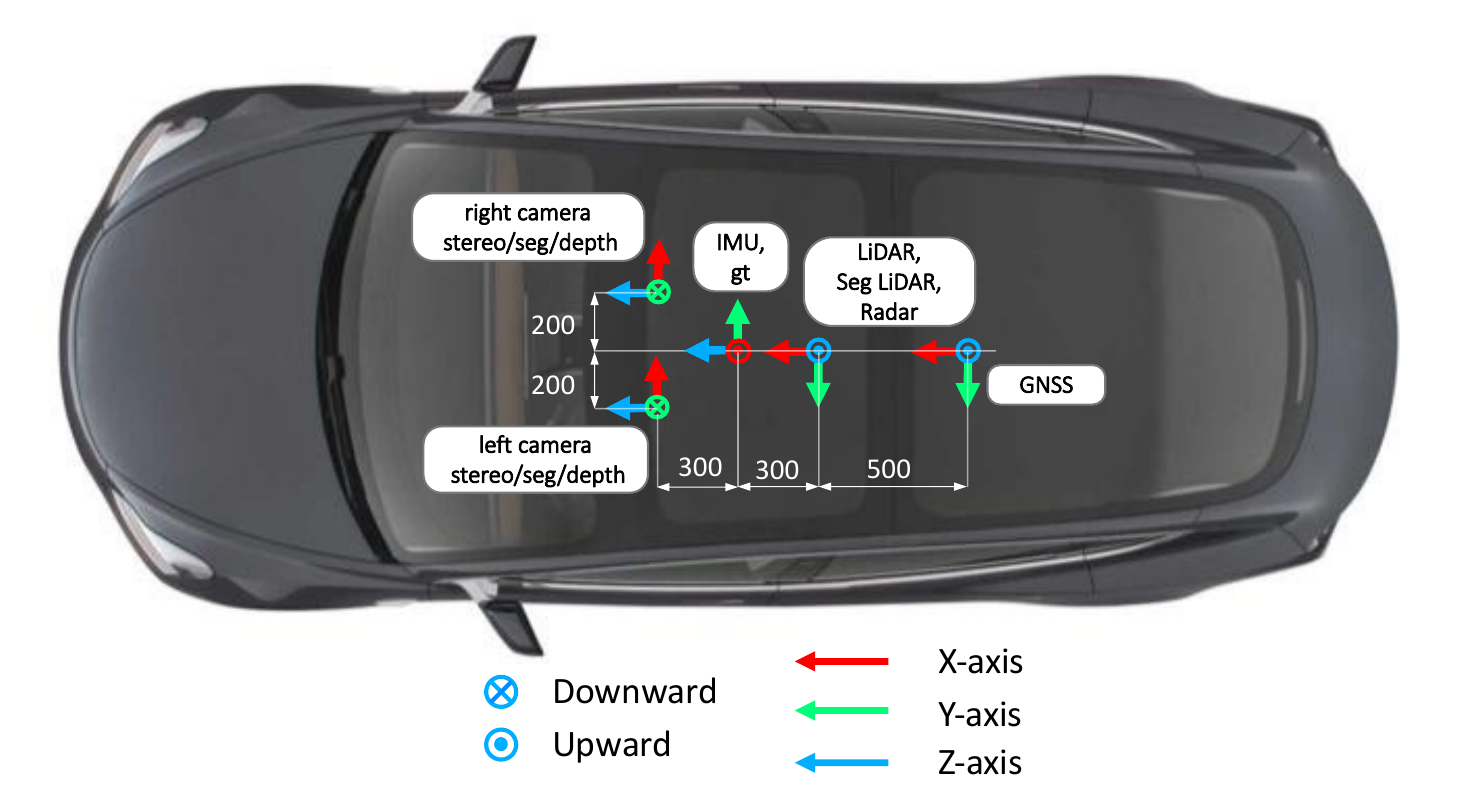}
    \caption{Sensor layout and coordinate systems of CARLA-Loc.}
    \label{fig:sensor layout}
\end{figure}

The layout of the sensor configuration is illustrated in Fig.~\ref{fig:sensor layout}. CARLA officially supports sensor simulation through its Blueprint Library. The callback function for each sensor is user-defined to process the received data. In CARLA-Loc, both sensor configurations and callback functions have been specifically designed to generate the final dataset.

\textbf{Visual sensors.} A binocular camera system is mounted on the front of the vehicle. This setup collects several types of visual data: RGB images, depth images, semantic segmentation images, and event camera data. The baseline distance between the cameras is 400 mm. Each camera features a 90$^\circ$ horizontal field of view (FOV) and captures images at a resolution of 1280$\times$720. The images are collected at a rate of 20 Hz, and the cameras are configured without any distortion.

\textbf{LiDAR.} 
The prototype of LiDAR is based on the Velodyne HDL-32E,\footnote{\href{https://velodynelidar.com/wp-content/uploads/2019/12/97-0038-Rev-N-97-0038-DATASHEETWEBHDL32E_Web.pdf}{https://velodynelidar.com/wp-content/uploads/2019/12/97-0038-Rev-N-97-0038-DATASHEETWEBHDL32E-Web.pdf}} a 32-beam spinning LiDAR capable of detecting objects up to 120 m away. It has an upper Field of View (FOV) of 10$^\circ$ and a lower FOV of -30$^\circ$ relative to the horizontal plane. LiDAR spins at a frequency of 20Hz, producing approximately 1,330k points per second. To more accurately replicate the distortion effect seen in point clouds, data from ten sectorial regions are collected separately during each full rotation and then combined into a single point cloud as one frame.

The official blueprints doesn't support adding the noise to point clouds of semantic LiDAR. Since 2 LiDARs (standard and semantic) in CARLA-Loc are located exactly the same. We collected ideal data from both LiDARs without noise and manually add Gaussian noise to them with the same random seed. The standard deviation is set to 0.02~m according to the real-world testing~\cite{9142208}.

\textbf{Radar.} The template of Radar is Oculii Falcon 77\footnote{\href{https://www.oculii.com/falcon}{https://www.oculii.com/falcon}}, which has 200m detection range, 120$^\circ$ azimuth FOV and 30$^\circ$ vertical FOV. Radar data is collected at 20Hz. 3 Radars are assembled to cover the full 360$^\circ$ field around the ego vehicle, which will return around 15,000 points in each second.

\textbf{IMU.} The IMU sensor receives significant optimization attention. It has been found that IMU data from the simulation environment is derived through direct numerical differentiation of the transformation. This approach leads to noticeable oscillations in IMU data during rapid vehicle starts and stops, which diverges from observations in real-world IMU data. To remedy this, cubic spline interpolation is applied to the original velocity and angular data. The derivatives at the sampling times are then computed to produce IMU data that is more realistic.

To introduce IMU noise, the following standard IMU measurement model is used:
\begin{equation}
    \Tilde{{a}}(t)={a}(t)+{b}_{a}(t)+{n}_{a}(t)
\end{equation}
\begin{equation}
    \Tilde{{\omega}}(t)={\omega}(t)+{b}_{\omega}(t)+{n}_{\omega}(t)
\end{equation}

${n(t)}$ is the white noise which is a zero-mean Gaussian noise process of strength $\sigma_n$, meaning $E[n(t)] = 0$ and $E[n(t_1)n(t_2)]={\sigma_n}^2 \delta(t_1-t_2)$. The whilte noise in discret time can be simulted as:
\begin{equation}
\label{eq.white_noise}
    n_d[i] = \sigma_{nd}w[i],
\end{equation}
where $w[i]\sim \mathcal{N}(0,1)$ and $\sigma_{nd} = \sigma \frac{1}{\sqrt{\Delta t}}$.

$b$ is the random walk,  with its time-inverse following a Gaussian process: $\Dot{b}(t)=\sigma_b$. The discrete-time case can be simulated as:
\begin{equation}
\label{eq.random_walk}
    b_d[i] = b_d[i-1] + \sigma_{b_d}w[i],
\end{equation}
where $w[i] \sim \mathcal{N}(0,1)$ and $\sigma_{bd}=\sigma_b \sqrt{\Delta t}$.

White noise of the acceleration and angular velocity in discrete-time is set to $2\times10^{-3} \ \mathrm{m/(s^2\cdot Hz)}$ and $2\times10^{-4} \ \mathrm{rad/(s\cdot Hz)}$, respectively, and their random walk is set to $3\times10^{-3} \ \mathrm{m/(s^3\cdot Hz)}$ and $2\times10^{-5} \ \mathrm{rad/(s^2\cdot Hz)}$, respectively. The discrete-time noise is incorporated into the IMU data by applying applying Eq.~\ref{eq.white_noise} and Eq.~\ref{eq.random_walk}. Additionally, the gravitational acceleration is also included to form final acceleration readings.

\textbf{GNSS.} The GNSS data is collected at 5 Hz. The standard deviation of longitude and latitude is set to be 3 m. The standard deviation of altitude is set to be 5 m.

\section{Experiment}

To validate the effectiveness of CARLA-Loc, visual-based SLAM methods and LiDAR-based SLAM methods are tested respectively on the proposed dataset. The computer used for experiment is equipped with an AMD Ryzen threadripper 1920x 12-core processor, an NVIDIA RTX 3090 graphic card and 64 GB of RAM. The loop closure detection functions (if any) have been deactivated for setup consistency.

The metric evaluated in the experiment is the Absolute Position Error (APE) and Relative Position Error (RPE), which is defined as:
\begin{equation}
\text {APE(RMSE)}=\sqrt{\frac{1}{N} \sum_{i=1}^N\left\|p_{\text {est}}(i)-p_{\text {gt}}(i)\right\|^2},
\end{equation}

\begin{equation}
\text {RPE(RMSE)}=\sqrt{\frac{1}{M}\sum_{i=1}^M \left\| \Delta p_{\text {est}}(i) - \Delta p_{\text {gt}}(i) \right\|^2},
\end{equation}
where $p_{\text {est }}(t)$ and $p_{\text {gt }}(t)$ represent the estimated and ground-truth positions at time $t$, respectively.  The relative position, $\Delta p$, is calculated as the difference between positions at two consecutive timestamps.

\subsection{VO and VIO experiment}

In this section, the performance of ORB-SLAM3~\cite{9440682}, VINS-Fusion~\cite{qin2019a}, and S-MSCKF~\cite{8258858} as visual (inertial) SLAM methods is evaluated. The experiments focus exclusively on stereo visual (SVO) or stereo visual-inertial (SVIO) configurations, deliberately avoiding monocular setups. The rationale behind this choice is the aim to specifically assess the influence of external factors on systems capable of accurately establishing scale. Since monocular VIO methods require scale initialization and employ varying strategies, they could introduce additional variables that might complicate the assessment.




\begin{table}[t]
\caption{Absolute Position Error (APE) of Visual SLAM Experiment}
\label{tab:visual_result}
\centering
{\fontsize{7}{8}\selectfont 
\begin{threeparttable}
\begin{tabular}{|c|c|ccccc|}
\hline
\multirow{3}{*}{\begin{tabular}[c]{@{}c@{}}map\\ sequence\end{tabular}} &
  \multirow{3}{*}{\begin{tabular}[c]{@{}c@{}}environment\\ setup$^*$\end{tabular}} &
  \multicolumn{5}{c|}{method} \\ \cline{3-7} 
 &
   &
  \multicolumn{1}{c|}{\begin{tabular}[c]{@{}c@{}}ORB3\\ SVO\end{tabular}} &
  \multicolumn{1}{c|}{\begin{tabular}[c]{@{}c@{}}ORB3\\ SVIO\end{tabular}} &
  \multicolumn{1}{c|}{\begin{tabular}[c]{@{}c@{}}VINS\\ SVO\end{tabular}} &
  \multicolumn{1}{c|}{\begin{tabular}[c]{@{}c@{}}VINS\\ SVIO\end{tabular}} &
  \begin{tabular}[c]{@{}c@{}}Stereo\\ MSCKF\end{tabular} \\ \hline \hline
\multirow{6}{*}{01} &
  CN(s) &
  \multicolumn{1}{c|}{7.15} &
  \multicolumn{1}{c|}{\textbf{0.53}} &
  \multicolumn{1}{c|}{4.62} &
  \multicolumn{1}{c|}{4.61} &
  5.45 \\ \cline{2-7} 
 &
  CN(d) &
  \multicolumn{1}{c|}{\textbf{4.30}} &
  \multicolumn{1}{c|}{32.25} &
  \multicolumn{1}{c|}{4.50} &
  \multicolumn{1}{c|}{4.57} &
  4.93 \\ \cline{2-7} 
 &
  FN(s) &
  \multicolumn{1}{c|}{\textbf{0.45}} &
  \multicolumn{1}{c|}{0.61} &
  \multicolumn{1}{c|}{4.79} &
  \multicolumn{1}{c|}{4.73} &
  3.81 \\ \cline{2-7} 
 &
  FN(d) &
  \multicolumn{1}{c|}{\textbf{3.51}} &
  \multicolumn{1}{c|}{3.87} &
  \multicolumn{1}{c|}{4.90} &
  \multicolumn{1}{c|}{4.79} &
  6.52 \\ \cline{2-7} 
 &
  RN(s) &
  \multicolumn{1}{c|}{0.94} &
  \multicolumn{1}{c|}{\textbf{0.72}} &
  \multicolumn{1}{c|}{4.52} &
  \multicolumn{1}{c|}{4.63} &
  4.26 \\ \cline{2-7} 
 &
  RN(d) &
  \multicolumn{1}{c|}{3.18} &
  \multicolumn{1}{c|}{\textbf{1.57}} &
  \multicolumn{1}{c|}{5.44} &
  \multicolumn{1}{c|}{6.36} &
  5.46 \\ \hline
\multirow{6}{*}{02} &
  CN(s) &
  \multicolumn{1}{c|}{fail} &
  \multicolumn{1}{c|}{\textbf{1.11}} &
  \multicolumn{1}{c|}{7.14} &
  \multicolumn{1}{c|}{11.95} &
  7.59 \\ \cline{2-7} 
 &
  CN(d) &
  \multicolumn{1}{c|}{8.32} &
  \multicolumn{1}{c|}{\textbf{1.61}} &
  \multicolumn{1}{c|}{7.96} &
  \multicolumn{1}{c|}{19.39} &
  8.61 \\ \cline{2-7} 
 &
  FN(s) &
  \multicolumn{1}{c|}{fail} &
  \multicolumn{1}{c|}{\textbf{2.53}} &
  \multicolumn{1}{c|}{8.21} &
  \multicolumn{1}{c|}{54.50} &
  7.88 \\ \cline{2-7} 
 &
  FN(d) &
  \multicolumn{1}{c|}{33.93} &
  \multicolumn{1}{c|}{124.99} &
  \multicolumn{1}{c|}{61.84} &
  \multicolumn{1}{c|}{52.27} &
  \textbf{7.88} \\ \cline{2-7} 
 &
  RN(s) &
  \multicolumn{1}{c|}{3.21} &
  \multicolumn{1}{c|}{\textbf{2.56}} &
  \multicolumn{1}{c|}{7.00} &
  \multicolumn{1}{c|}{6.90} &
  6.91 \\ \cline{2-7} 
 &
  RN(d) &
  \multicolumn{1}{c|}{23.38} &
  \multicolumn{1}{c|}{\textbf{1.83}} &
  \multicolumn{1}{c|}{10.82} &
  \multicolumn{1}{c|}{26.26} &
  8.27 \\ \hline
\multirow{6}{*}{03} &
  CN(s) &
  \multicolumn{1}{c|}{58.56} &
  \multicolumn{1}{c|}{\textbf{4.09}} &
  \multicolumn{1}{c|}{12.50} &
  \multicolumn{1}{c|}{92.82} &
  fail \\ \cline{2-7} 
 &
  CN(d) &
  \multicolumn{1}{c|}{146.36} &
  \multicolumn{1}{c|}{fail} &
  \multicolumn{1}{c|}{\textbf{24.28}} &
  \multicolumn{1}{c|}{fail} &
  fail \\ \cline{2-7} 
 &
  FN(s) &
  \multicolumn{1}{c|}{22.21} &
  \multicolumn{1}{c|}{\textbf{3.73}} &
  \multicolumn{1}{c|}{118.65} &
  \multicolumn{1}{c|}{fail} &
  23.01 \\ \cline{2-7} 
 &
  FN(d) &
  \multicolumn{1}{c|}{\textbf{159.24}} &
  \multicolumn{1}{c|}{fail} &
  \multicolumn{1}{c|}{213.56} &
  \multicolumn{1}{c|}{fail} &
  fail \\ \cline{2-7} 
 &
  RN(s) &
  \multicolumn{1}{c|}{\textbf{31.53}} &
  \multicolumn{1}{c|}{39.30} &
  \multicolumn{1}{c|}{343.85} &
  \multicolumn{1}{c|}{fail} &
  68.05 \\ \cline{2-7} 
 &
  RN(d) &
  \multicolumn{1}{c|}{171.77} &
  \multicolumn{1}{c|}{fail} &
  \multicolumn{1}{c|}{fail} &
  \multicolumn{1}{c|}{fail} &
  fail \\ \hline
\multirow{6}{*}{04} &
  CN(s) &
  \multicolumn{1}{c|}{\textbf{2.81}} &
  \multicolumn{1}{c|}{6.65} &
  \multicolumn{1}{c|}{5.17} &
  \multicolumn{1}{c|}{4.94} &
  3.54 \\ \cline{2-7} 
 &
  CN(d) &
  \multicolumn{1}{c|}{10.07} &
  \multicolumn{1}{c|}{4.67} &
  \multicolumn{1}{c|}{5.17} &
  \multicolumn{1}{c|}{5.41} &
  \textbf{3.82} \\ \cline{2-7} 
 &
  FN(s) &
  \multicolumn{1}{c|}{\textbf{0.27}} &
  \multicolumn{1}{c|}{5.70} &
  \multicolumn{1}{c|}{4.88} &
  \multicolumn{1}{c|}{36.03} &
  3.58 \\ \cline{2-7} 
 &
  FN(d) &
  \multicolumn{1}{c|}{60.75} &
  \multicolumn{1}{c|}{9.50} &
  \multicolumn{1}{c|}{14.62} &
  \multicolumn{1}{c|}{49.59} &
  \textbf{3.37} \\ \cline{2-7} 
 &
  RN(s) &
  \multicolumn{1}{c|}{\textbf{0.60}} &
  \multicolumn{1}{c|}{5.85} &
  \multicolumn{1}{c|}{5.16} &
  \multicolumn{1}{c|}{10.84} &
  3.16 \\ \cline{2-7} 
 &
  RN(d) &
  \multicolumn{1}{c|}{26.65} &
  \multicolumn{1}{c|}{6.56} &
  \multicolumn{1}{c|}{fail} &
  \multicolumn{1}{c|}{\textbf{5.44}} &
  fail \\ \hline
\multirow{6}{*}{05} &
  CN(s) &
  \multicolumn{1}{c|}{3.15} &
  \multicolumn{1}{c|}{3.24} &
  \multicolumn{1}{c|}{3.01} &
  \multicolumn{1}{c|}{4.03} &
  \textbf{1.93} \\ \cline{2-7} 
 &
  CN(d) &
  \multicolumn{1}{c|}{76.40} &
  \multicolumn{1}{c|}{\textbf{2.29}} &
  \multicolumn{1}{c|}{26.29} &
  \multicolumn{1}{c|}{3.97} &
  fail \\ \cline{2-7} 
 &
  FN(s) &
  \multicolumn{1}{c|}{\textbf{0.23}} &
  \multicolumn{1}{c|}{23.52} &
  \multicolumn{1}{c|}{4.02} &
  \multicolumn{1}{c|}{fail} &
  5.05 \\ \cline{2-7} 
 &
  FN(d) &
  \multicolumn{1}{c|}{80.99} &
  \multicolumn{1}{c|}{555.48} &
  \multicolumn{1}{c|}{\textbf{72.67}} &
  \multicolumn{1}{c|}{fail} &
  fail \\ \cline{2-7} 
 &
  RN(s) &
  \multicolumn{1}{c|}{\textbf{0.28}} &
  \multicolumn{1}{c|}{18.03} &
  \multicolumn{1}{c|}{3.18} &
  \multicolumn{1}{c|}{fail} &
  fail \\ \cline{2-7} 
 &
  RN(d) &
  \multicolumn{1}{c|}{81.12} &
  \multicolumn{1}{c|}{425.74} &
  \multicolumn{1}{c|}{80.77} &
  \multicolumn{1}{c|}{\textbf{6.76}} &
  fail \\ \hline
\multirow{6}{*}{06} &
  CN(s) &
  \multicolumn{1}{c|}{fail} &
  \multicolumn{1}{c|}{20.65} &
  \multicolumn{1}{c|}{7.09} &
  \multicolumn{1}{c|}{11.33} &
  \textbf{6.41} \\ \cline{2-7} 
 &
  CN(d) &
  \multicolumn{1}{c|}{89.01} &
  \multicolumn{1}{c|}{\textbf{1.51}} &
  \multicolumn{1}{c|}{6.44} &
  \multicolumn{1}{c|}{fail} &
  6.62 \\ \cline{2-7} 
 &
  FN(s) &
  \multicolumn{1}{c|}{\textbf{0.51}} &
  \multicolumn{1}{c|}{1.82} &
  \multicolumn{1}{c|}{6.43} &
  \multicolumn{1}{c|}{fail} &
  6.44 \\ \cline{2-7} 
 &
  FN(d) &
  \multicolumn{1}{c|}{\textbf{0.96}} &
  \multicolumn{1}{c|}{fail} &
  \multicolumn{1}{c|}{6.81} &
  \multicolumn{1}{c|}{fail} &
  6.40 \\ \cline{2-7} 
 &
  RN(s) &
  \multicolumn{1}{c|}{2.23} &
  \multicolumn{1}{c|}{\textbf{2.12}} &
  \multicolumn{1}{c|}{fail} &
  \multicolumn{1}{c|}{fail} &
  6.35 \\ \cline{2-7} 
 &
  RN(d) &
  \multicolumn{1}{c|}{12.84} &
  \multicolumn{1}{c|}{106.95} &
  \multicolumn{1}{c|}{fail} &
  \multicolumn{1}{c|}{fail} &
  \textbf{6.77} \\ \hline
\multirow{6}{*}{07} &
  CN(s) &
  \multicolumn{1}{c|}{\textbf{1.65}} &
  \multicolumn{1}{c|}{1.97} &
  \multicolumn{1}{c|}{4.89} &
  \multicolumn{1}{c|}{4.62} &
  5.32 \\ \cline{2-7} 
 &
  CN(d) &
  \multicolumn{1}{c|}{4.69} &
  \multicolumn{1}{c|}{\textbf{1.92}} &
  \multicolumn{1}{c|}{4.95} &
  \multicolumn{1}{c|}{4.89} &
  5.85 \\ \cline{2-7} 
 &
  FN(s) &
  \multicolumn{1}{c|}{\textbf{0.90}} &
  \multicolumn{1}{c|}{2.01} &
  \multicolumn{1}{c|}{7.16} &
  \multicolumn{1}{c|}{fail} &
  8.35 \\ \cline{2-7} 
 &
  FN(d) &
  \multicolumn{1}{c|}{54.62} &
  \multicolumn{1}{c|}{\textbf{2.51}} &
  \multicolumn{1}{c|}{92.21} &
  \multicolumn{1}{c|}{17.42} &
  10.18 \\ \cline{2-7} 
 &
  RN(s) &
  \multicolumn{1}{c|}{1.90} &
  \multicolumn{1}{c|}{\textbf{1.61}} &
  \multicolumn{1}{c|}{4.82} &
  \multicolumn{1}{c|}{6.21} &
  6.70 \\ \cline{2-7} 
 &
  RN(d) &
  \multicolumn{1}{c|}{60.38} &
  \multicolumn{1}{c|}{\textbf{1.54}} &
  \multicolumn{1}{c|}{14.17} &
  \multicolumn{1}{c|}{5.30} &
  6.59 \\ \hline
\end{tabular}
\begin{tablenotes}
  \item[*] CN, FN, RN stand for Clear Noon, Foggy Noon, and Rainy Night, respectively.
  (s) and (d) indicate static or dynamic presets.
\end{tablenotes}
\end{threeparttable}
}
\end{table}

\begin{figure}[h]
    \includegraphics[width = 8.8cm]{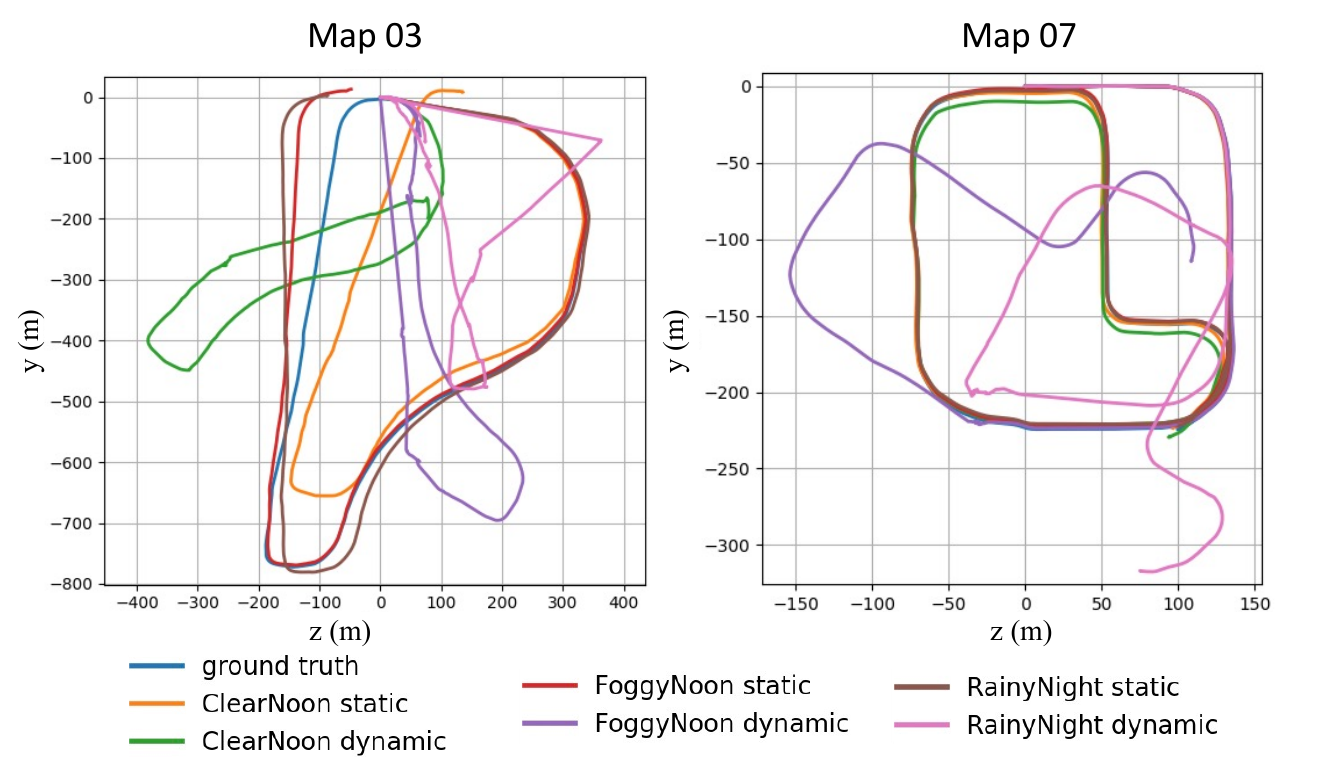}
    \caption{Trajectories of ORB-SLAM3 (stereo mode) in sequences from map 03 and map 07, the presence of dynamic object, as well as the poor weather condition evidently affect the overall localization performance.  }
    \label{fig:vslam_plot}
\end{figure}




The results of VO and VIO experiment is shown in Tab.~\ref{tab:visual_result}. ORB-SLAM3 outperforms the other two methods. Out of all 42 sequences, ORB-SLAM3 SVIO achieves the highest localization accuracy in 17 sequences, followed by the non-inertial mode of it with 14 sequences. When considering different presets of the same map, sequences with dynamic objects exhibit a significant decrease in accuracy. This is in line with our expectations, and to illustrate this difference, we specifically plot the trajectories of the ORB-SLAM3 SVO mode under map 03 and map 07 (Fig.~\ref{fig:vslam_plot}).
The inclusion of inertial sensors to some extent enhances the SLAM method's resistance to dynamic objects. Among all 21 dynamic sequences, ORB-SLAM3 SVIO mode has a superior accuracy in 11 sequences compared to the pure visual mode, while VINS SVIO outperforms the pure visual mode in 11 sequences.

Regarding the influence of weather factors, the intuition is that localization under rainy should be worse that clear weather. However, in static environments, many sequences under "Clear Noon" have poorer localization accuracy than "Foggy Noon". Root reason behind this could be that clear weather conditions increase the presence of distant feature points. The disparity errors of stereo pixels at these faraway points tend to be relatively large, which negatively affects the overall localization performance.
\subsection{LO and LIO experiment}

The LiDAR SLAM method selected to test are LOAM~\cite{zhang2014loam}, LeGo-LOAM~\cite{8594299}, and FAST-LIO2~\cite{9697912}. LOAM is the first feature-based 3D LiDAR SLAM proposed by Zhang et. al., while LeGo-LOAM further modifies the original pipeline to identify the ground and other features. FAST-LIO2 is a LiDAR-Interial SLAM method that builds the small batch assumption (every point belongs to a small plane), which fully leverages all point cloud in scans which out feature extraction. The sensor fusion is completed through iterated Kalman filter for state estimation.

\begin{table}[htb]
\caption{Absolute Position Error (APE) of LiDAR SLAM Experiment}
\label{tab:lidar_result}
\centering 
{\fontsize{7}{8}\selectfont 
\begin{threeparttable}
\begin{tabular}{|c|c|cccc|}
\hline
\multirow{3}{*}{\begin{tabular}[c]{@{}c@{}}map\\ sequence\end{tabular}} &
  \multicolumn{1}{c|}{\multirow{3}{*}{\begin{tabular}[c]{@{}c@{}}dynamic\\ level$^*$\end{tabular}}} &
  \multicolumn{4}{c|}{method} \\ \cline{3-6} 
 &
  \multicolumn{1}{c|}{} &
  \multicolumn{1}{c|}{ALOAM} &
  \multicolumn{1}{c|}{\begin{tabular}[c]{@{}c@{}}LEGO\\ LOAM\end{tabular}} &
  \multicolumn{1}{c|}{\begin{tabular}[c]{@{}c@{}}FAST\\ LIO2\end{tabular}} &
  \begin{tabular}[c]{@{}c@{}}FAST \\ LIO2(F)$^1$\end{tabular} \\ \hline \hline
\multirow{3}{*}{01} & raw (s) & \multicolumn{1}{c|}{\textbf{1.82}} & \multicolumn{1}{c|}{7.21}   & \multicolumn{1}{c|}{2.09}          & 2.06          \\ \cline{2-6} 
                    & seg (d) & \multicolumn{1}{c|}{1.86}          & \multicolumn{1}{c|}{7.21}   & \multicolumn{1}{c|}{1.83}          & \textbf{1.26} \\ \cline{2-6} 
                    & raw (d) & \multicolumn{1}{c|}{2.87}          & \multicolumn{1}{c|}{10.66}  & \multicolumn{1}{c|}{1.77}          & \textbf{1.15} \\ \hline
\multirow{3}{*}{02} & raw (s) & \multicolumn{1}{c|}{9.45}          & \multicolumn{1}{c|}{9.34}   & \multicolumn{1}{c|}{2.71}          & \textbf{2.39} \\ \cline{2-6} 
                    & seg (d) & \multicolumn{1}{c|}{9.23}          & \multicolumn{1}{c|}{5.96}   & \multicolumn{1}{c|}{2.77}          & \textbf{2.08} \\ \cline{2-6} 
                    & raw (d) & \multicolumn{1}{c|}{10.54}         & \multicolumn{1}{c|}{9.95}   & \multicolumn{1}{c|}{\textbf{2.43}} & 2.54          \\ \hline
\multirow{3}{*}{03} & raw (s) & \multicolumn{1}{c|}{35.66}         & \multicolumn{1}{c|}{308.96} & \multicolumn{1}{c|}{\textbf{4.02}} & 4.56          \\ \cline{2-6} 
                    & seg (d) & \multicolumn{1}{c|}{55.54}         & \multicolumn{1}{c|}{459.25} & \multicolumn{1}{c|}{\textbf{4.16}} & 17.41         \\ \cline{2-6} 
                    & raw (d) & \multicolumn{1}{c|}{341.11}        & \multicolumn{1}{c|}{441.48} & \multicolumn{1}{c|}{\textbf{4.71}} & 16.20         \\ \hline
\multirow{3}{*}{04} & raw (s) & \multicolumn{1}{c|}{2.13}          & \multicolumn{1}{c|}{9.59}   & \multicolumn{1}{c|}{\textbf{1.36}} & 6.64          \\ \cline{2-6} 
                    & seg (d) & \multicolumn{1}{c|}{\textbf{1.58}} & \multicolumn{1}{c|}{9.19}   & \multicolumn{1}{c|}{1.83}          & 13.07         \\ \cline{2-6} 
                    & raw (d) & \multicolumn{1}{c|}{\textbf{1.68}} & \multicolumn{1}{c|}{9.61}   & \multicolumn{1}{c|}{1.86}          & 9.08          \\ \hline
\multirow{3}{*}{05} & raw (s) & \multicolumn{1}{c|}{4.53}          & \multicolumn{1}{c|}{33.45}  & \multicolumn{1}{c|}{2.36}          & \textbf{0.84} \\ \cline{2-6} 
                    & seg (d) & \multicolumn{1}{c|}{23.70}         & \multicolumn{1}{c|}{56.40}  & \multicolumn{1}{c|}{2.76}          & \textbf{1.83} \\ \cline{2-6} 
                    & raw (d) & \multicolumn{1}{c|}{93.64}         & \multicolumn{1}{c|}{141.78} & \multicolumn{1}{c|}{2.70}          & \textbf{1.98} \\ \hline
\multirow{3}{*}{06} & raw (s) & \multicolumn{1}{c|}{\textbf{2.40}} & \multicolumn{1}{c|}{52.19}  & \multicolumn{1}{c|}{3.01}          & 2.70          \\ \cline{2-6} 
                    & seg (d) & \multicolumn{1}{c|}{\textbf{1.95}} & \multicolumn{1}{c|}{46.56}  & \multicolumn{1}{c|}{2.89}          & 2.76          \\ \cline{2-6} 
                    & raw (d) & \multicolumn{1}{c|}{\textbf{1.95}} & \multicolumn{1}{c|}{101.89} & \multicolumn{1}{c|}{2.88}          & 7.75          \\ \hline
\multirow{3}{*}{07} & raw (s) & \multicolumn{1}{c|}{5.52}          & \multicolumn{1}{c|}{23.37}  & \multicolumn{1}{c|}{12.02}         & \textbf{3.78} \\ \cline{2-6} 
                    & seg (d) & \multicolumn{1}{c|}{5.97}          & \multicolumn{1}{c|}{37.09}  & \multicolumn{1}{c|}{\textbf{4.47}} & 11.88         \\ \cline{2-6} 
                    & raw (d) & \multicolumn{1}{c|}{16.72}         & \multicolumn{1}{c|}{147.27} & \multicolumn{1}{c|}{2.75}          & \textbf{2.59} \\ \hline
\hline
\end{tabular}
\begin{tablenotes}
  \item[*] ‘raw’ stands for raw point cloud data, (s) and (d) indicate the scenario is static or dynamic. ‘seg (d)’ stands for removing the points belong to objects through semantic segmentation label from the raw dynamic sequence.
  
  \item[1] (F) stands for enabling future extraction. Otherwise, use small batch assumption and search all point-to-plane residual. This option is set off in FAST-LIO by default.
\end{tablenotes}
\end{threeparttable}
}
\end{table}

\begin{figure}[htb]
    \centerline{\includegraphics[width = 8.8cm]{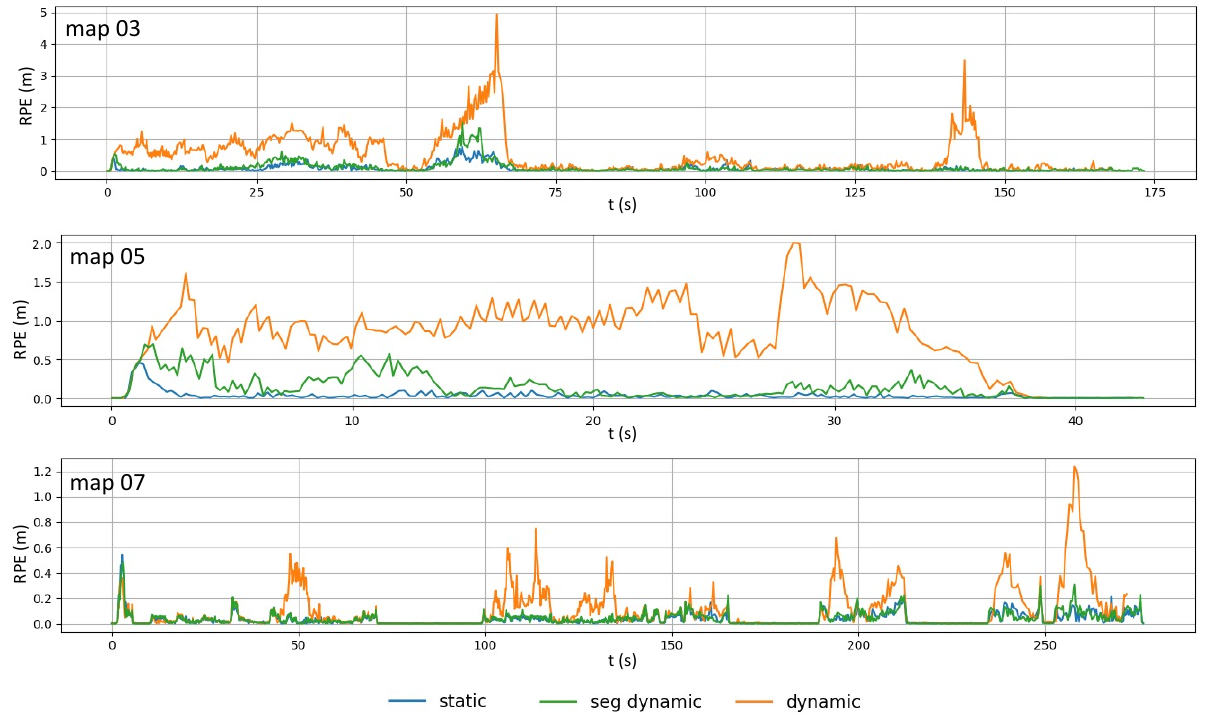}}
    \caption{Relative Position Error of A-LOAM in map 02, map 05 and map 07. In the presence of dynamic objects (orange line), the estimation error of relative positions is evidently greater. However, this impact can be mitigated using semantic segmentation label to exclude dynamic points (green line).}
    \label{fig:aloam_rpe}
\end{figure}


The experiment result is shown in Tab.~\ref{tab:lidar_result}.

In methods other than FAST-LIO2, the presence of dynamic objects typically decrease localization accuracy. Due to the integration of inertial sensors, FAST-LIO2(F) exhibits notably lower errors in dynamic scenes compared to ALOAM and LeGO-LOAM.

Using feature extraction also poses impact on the final localization precision. The vanilla FAST-LIO2, demonstrates impressive robustness. The presence of dynamic objects hardly affect the localization accuracy. A-LOAM and LeGO-LOAM show accuracy loss in dynamic scenes, especially evident in maps 03, 04, and 06. Additionally, LeGO-LOAM, aiming for lightweight performance and using only segmented above ground features, is the most susceptible to dynamic objects in our tests.

Regarding sequences preprocessed using semantic segmentation, the localization accuracy typically falls between static and dynamic. Result of A-LOAM is given in Fig.~\ref{fig:aloam_rpe} for example. We believe that since the hole areas from removed object points are directly skipped, leading to imprecise feature extraction for points at the edges. However, the error is still obviously smaller than raw dynamic sequence.

\section{Conclusion}

In this paper, we propose CARLA-loc, simulation-generated dataset with well-configured sensor configuration. The map selection, environmental settings and sensor configurations are all well-designed to make the data divers and realistic. The script is open-sourced for customization and we also provide user-friendly ros bag creation toolkit.

In the experiment section, both visual-based and LiDAR-based SLAM show improvements in localization accuracy with the integration of IMU, proving the robustness enhancement of multi-sensor SLAM. The accuracy generally decreases when the environment becomes more challenging. An exception occurs in the Foggy Noon scenario where only nearby area is visible, and the accuracy unexpectedly increases. We believe this is due to the larger parallax of points near the camera, which results in more accurate depth estimation, essentially equivalent to adding the weight of reliable features. A related discussion on this can be found in ~\cite{9981694}. The LiDAR experiment demonstrates the advantages of using direct matching over curvature-based feature extraction like LOAM, especially in highly dynamic scenarios.

Improving the robustness and stability of SLAM algorithms still remains challenging. The proposed dataset is believed to provide a better testing and evaluation benchmark for the SLAM community.

\newpage




\bibliographystyle{ieeetr}
\bibliography{reference}

\end{document}